# MACHINE LEARNING-BASED SOFT SENSORS FOR VACUUM DISTILLATION UNIT

*Kamil Oster, Department of Mathematics, The University of Manchester, UK; Process Integration Limited, UK*

*Stefan Güttel, Department of Mathematics, The University of Manchester, UK*

*Lu Chen, Process Integration Limited, UK*

*Jonathan Shapiro, Department of Computer Science, The University of Manchester, UK*

*Megan Jobson, Department of Chemical Engineering and Analytical Science, UK*

**Introduction**

Product quality assessment in the petroleum processing industry, such as crude distillation, can be difficult and time-consuming, *e.g.* due to a manual collection of liquid samples from the plant and the subsequent chemical laboratory analysis of the samples. The product quality is an important property that informs whether the products of the process are within the regulated specifications, such as ASTM Petroleum Standards. In particular, the delays caused by sample processing (collection, laboratory measurements, results analysis, reporting, *etc.*) can lead to detrimental economic effects. One of the strategies to deal with this problem is so-called soft sensors.

Soft sensors are a collection of models that can be used to predict and forecast some infrequently measured properties (such as laboratory measurements of petroleum products) based on more frequent measurements of quantities like temperature, pressure and flow rate provided by physical sensors [1]. Soft sensors short-cut the pathway to obtain relevant information about the product quality, often providing relevant measurements as frequently as every minute. One of the applications of soft sensors is for the real-time optimization of a chemical process by a targeted adaptation of operating parameters. Models used for soft sensors can have various forms, however, among the most common are those based on artificial neural networks (ANNs) [2].

While soft sensors can deal with some of the issues in the refinery processes, their development and deployment can pose other challenges that are addressed in this paper. Firstly, it is important to enhance the quality of both sets of data (laboratory measurements and physical sensors) in a so-called data pre-processing stage (as described in Methodology section) [3]. Secondly, once the data sets are pre-processed, different models need to be tested against prediction error and the model's interpretability. In this work, we present a framework for soft sensor development from raw data to ready-to-use models.

**Methodology**

The example process data used in this paper are taken from a vacuum distillation unit (VDU) of a refinery. All data are taken from an Asian refinery. The VDU, and all physical sensors and product lines (side draws), is shown in **Figure 1**. The first category of data is measurements of temperature ($T$), pressure ($p$) and flow rate ($F$) from physical sensors that make up the input data for the soft sensors. The data set consists of 40 physical sensors. 31 physical sensors were taken for further analysis, with the others rejected due to a high noise level and a large amount of missing values (sensors F20, F10, F30, F41), and coking causing problems with the sensors on the 6$^{th}$ side draw, V6SS (sensors T6, T61, F61, F62, F52).

The second category of data is ASTM-D2887 laboratory measurements of the boiling curve of a distillation product of the 3$^{rd}$ side draw vacuum distillation product, namely V3SS.

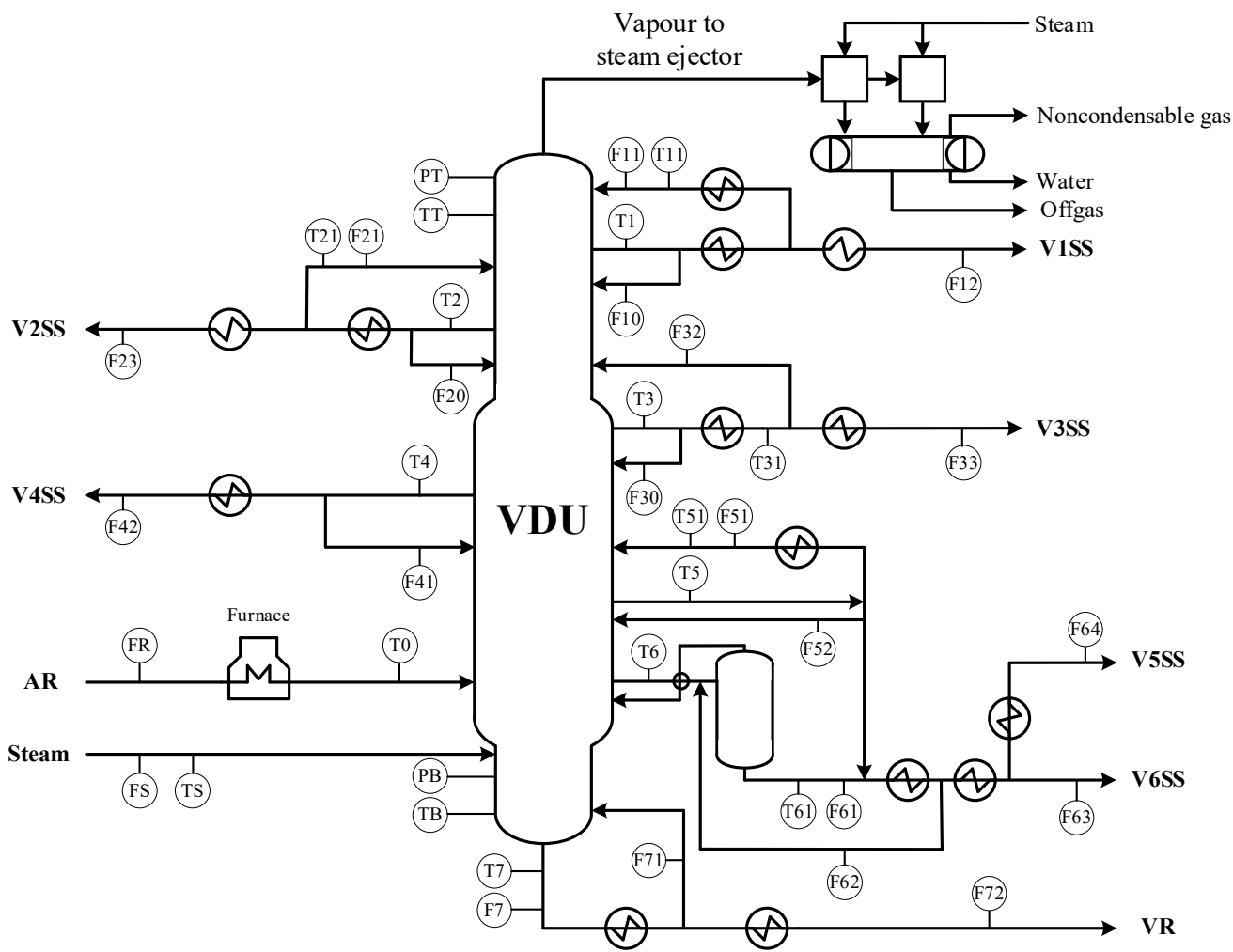

**Figure 1.** VDU diagram with all physical sensors (*T* - temperature, *p* - pressure, *F* - flow rate) and product lines (V1SS, V2SS, V3SS, V4SS, V5SS, V6SS, VR - vacuum residue), as well as inlets (AR - atmospheric residue).

The quality of the products of V1SS, V2SS, V4SS, V5SS, V6SS and VR is measured only once a week, which significantly limits the amount of data available to develop soft sensors. Hence, in this work, we only focused on V3SS that is measured daily. The developed soft sensors use measured data of physical sensors to infer points on the distillation curve (namely 2, 10, 30, 50, 70, 100 vol%), later called distillation points. The main aim of our framework is to ensure robustness, reliability, reproducibility and interpretability of the models. The overall framework (steps and data flow) is presented in **Figure 2.**

     Firstly, the data from the laboratory measurements undergo pre-processing which includes removal of zero/missing/duplicated values and outliers. For the outlier detection, the interquartile method (IQR) was used [4]. For each distillation point, upper and lower bounds were determined and used to identify data points that exceeded the calculated IQR thresholds. The data for each distillation point were combined to establish data at which V3SS was outside the average behavior for at least one of the distillation points.

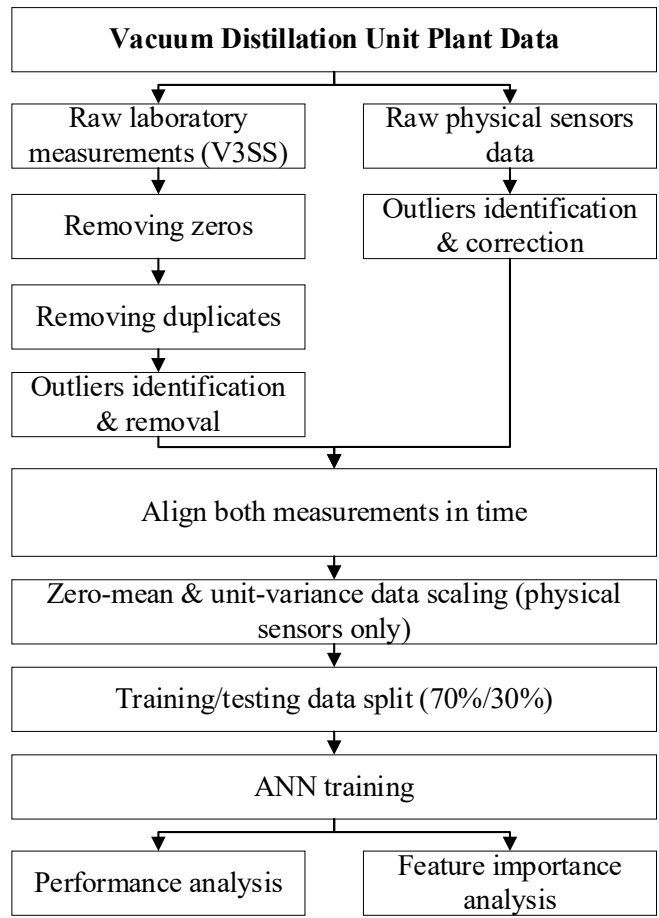

**Figure 2.** Steps of the framework and data flow.

A main issue in developing soft sensors is the assessment of the model's performance. For new models, it is important to understand how well the model can perform for the data used, specifically what prediction error threshold may be acceptable. One of the approaches to ascertain these thresholds is to use some established models. The V3SS measurements can be treated as a time series, therefore, methods for time series prediction can be employed [5].

We used one of the most commonly used time series models – Seasonal AutoRegressive Integrated Moving Average (SARIMA) [6,7]. The seasonality component in SARIMA is expected to originate from crude oil supply changes and work shift patterns. The *pmdarima* Python package was used to grid search the best SARIMA model for all V3SS distillation points. The algorithm aims to minimize the Akaike Information Criterion (AIC) in a stepwise manner to find the best model for the hyperparameters (using the Dickey-Fuller test of stationarity and the Canova-Hansen test for seasonality). The details of this approach can be found in the original documentation for the package [8]. Once the best SARIMA models for each distillation point were found, these models are then used to predict the distillation points data. Mean absolute errors between predicted (through SARIMA) and actual data for each distillation point are used as upper thresholds for the new ANN model's prediction errors.

In our previous work [9], we studied outlier detection in the plant sensor data ($T$, $p$ and $F$). Generally, two types of outliers exist: short-term outliers (*i.e.* noise and erroneous data) and long-term outliers (*i.e.* malfunction for a longer period of time). The outlier detection is a two-step process: 1) identification and removal of short-term outliers; 2) identification of long-term outlier periods. Briefly, change points are identified across each of the signals and then used to partition the signals. The change point detection method is based on maximum log-likelihood estimation (assuming changes in the average response of the signal), ensuring that the partitioned signals comply with the central limit theorem, creating piecewise constant approximations of the mean and standard deviation. These piecewise approximated standard deviations are used to calculate upper and lower thresholds for the outlier detection (specifically triple standard deviation values - $3\sigma$ method), and piecewise $3\sigma$ upper and lower thresholds. The short-term outliers identified with this method were replaced with the piecewise constant means before proceeding to identify long-term outliers. As discussed in [9], principal component analysis (PCA) combined with Hotteling's $T^2$ statistics was used to identify the long-term outlier periods. The V3SS data during long-term outlier periods were discarded from further analysis.

The next step in the framework is the time alignment of both data sets, as the V3SS data were recorded daily, while plant data ($p$, $T$, $F$) were recorded every minute. To capture the state of the VDU process immediately prior to withdrawing a sample for laboratory measurement, a 1-hour average was calculated before each of the laboratory measurements. The pre-processed and time-aligned input and output data were scaled appropriately (with zero-mean and unit-variance). Afterwards, the data were split into training and testing sets (70% and 30%, respectively).

Our soft sensor model is based on a multilayer perceptron neural network. The input layer for the model receives 1-hour average plant data (with 31 features/sensors in total), while the output layer predicts the V3SS laboratory measurements (each record is an array of 7 distillation points, *i.e.* from 2 vol% to 100 vol%). The tested ANN model contains 2 hidden layers, each with 30 nodes. The input layer and all hidden layers use ReLu activation functions. The weights and biases of each hidden layer were initialized using the Xavier normal initializer, and zeros, respectively. The Keras/TensorFlow package was used to implement and train the neural networks. The neural network was optimized for a Huber loss function and with the Adam optimizer (both with default parameters from Keras/TensorFlow). Model checkpoint callback was used to prevent overfitting: after each epoch, the algorithm calls back to calculate the prediction error on the testing set to evaluate whether the model improved in relation to the testing set. In this approach, the algorithm only saves the model that showed improvement on the testing set, while updating the weights and biases on training set. The model training ran for up to 10,000 epochs.

To understand how all features correlate with the predicted output through the trained model, feature importance was analyzed. SHapley Additive exPlanations (SHAP) was used for this purpose [10]. Briefly, SHAP is based on calculating Shapley values as in game theory for each of the features (sensors). Each feature (sensor) value of the instance is a 'player' in a game, while the prediction is the 'payout'. Shapley values inform about fair distribution of the 'payout' amongst the features. Aside from minimizing the prediction error, it is important to ensure the model's interpretability, *e.g.* through feature importance analysis, to ensure that the impact of features - plant physical sensors on the predicted output is reflected in an overall understanding of the underlying physicochemical behavior of the VDU.

**Results and Discussion**

The laboratory measurements used in this paper follow the industrial standard ASTM D2887 *Standard Test Method for Distillation of Petroleum Products at Reduced Pressure* [11]. These measurements are carried out on a liquid sample withdrawn from the VDU to represent the product quality. In this paper, a product of 3rd side draw of VDU is studied, *i.e.* light-vacuum gas oil, namely V3SS.

The frequency of the measurements is 1 per day. Overall, the raw dataset consists of 6,037 data points. **Figure 3a** shows all 7 distillation points of the V3SS distillation curve - 2 vol%, 10 vol%, 30 vol%, 50 vol%, 70 vol%, 90 vol% and 100 vol%. Upon visual inspection of the raw data, it can be seen that a large number of data points are nulls (probable artifacts in the laboratory information management system). Moreover, duplicated records were found in the data. Overall, the raw data consisted of 4,391 nulls (72.7% of all data) and 347 duplicates (5.7% of all data). After removing these artifacts, the remaining dataset was 1,299 data points (shown in **Figure 3b**). In the next step, outliers were detected and removed (in accordance with IQR method described in Methodology section), as shown in **Figure 3c.** In total, 91 outliers were detected and removed which leaves 1,208 data points. Interestingly, there were periods that would suggest issues with the functioning of the VDU system (35 outliers in the entire period between days 395 and 557). These time instances may correspond to crude oil supply changes that can lead to such deviations in product quality.

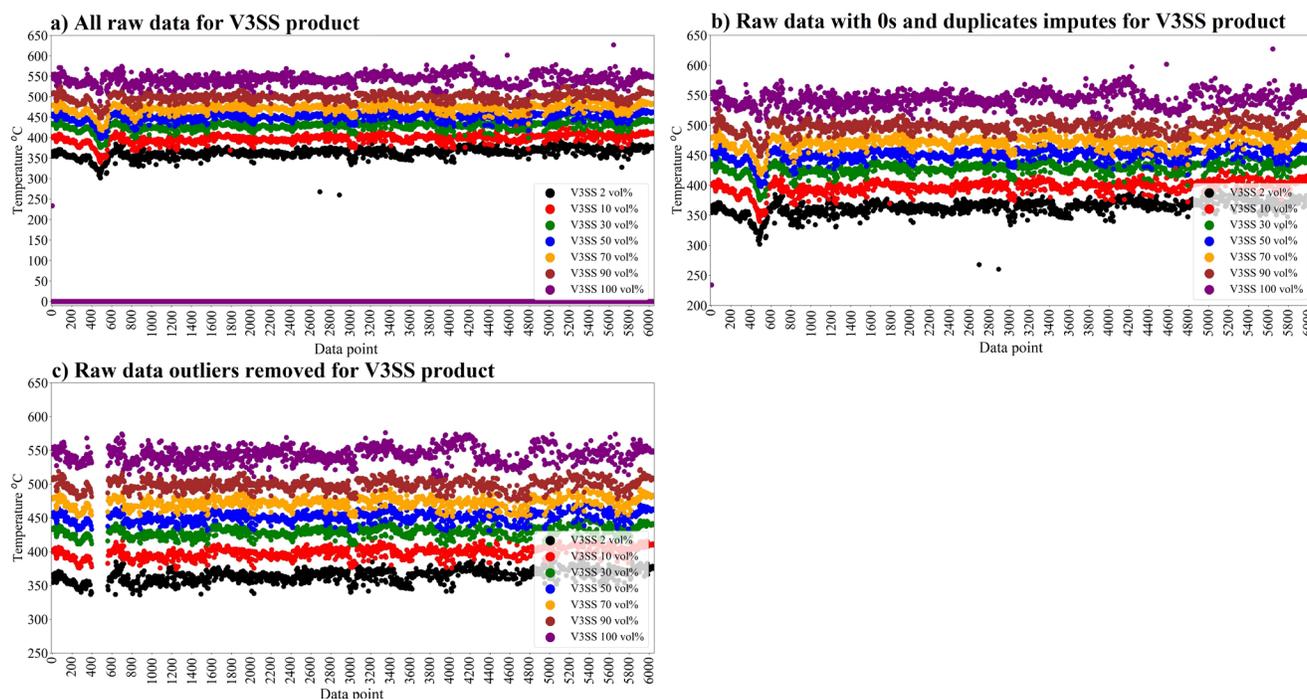

**Figure 3.** Output data (laboratory measurements of V3SS distillation product) pre-processing: **a)** original raw data; **b)** data after removing nulls and duplicated records; **c)** data after removing outliers.

As described in the Methodology section, SARIMA modelling was used to ascertain the V3SS accuracy, as a baseline for the predictive performance of the ANN. The results of the predicted data against the actual data for V3SS can be seen in **Figure 4a.** The values in the last column in the table (**Figure 4b**) will be used in a later part of this paper to assess the performance of the ANN, *i.e.* the maximum average error for the predictions for each distillation point. For example, it is expected that the ANN is able to predict V3SS 100 vol% with an error below 7.7°C.

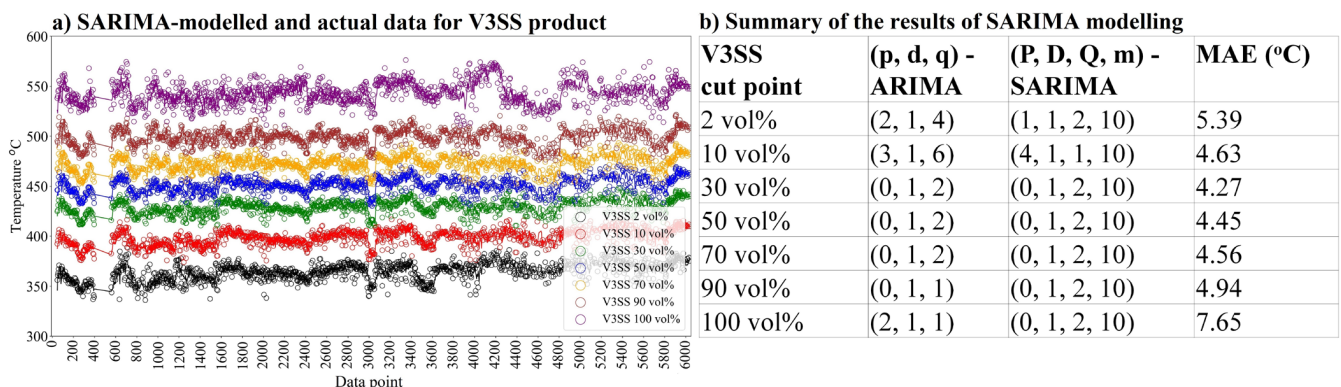

**Figure 4. a)** Comparison of SARIMA-modelled and actual data for V3SS distillation points; **b)** coefficients for ARIMA and SARIMA components with mean absolute error (MAE) of the deviations.

The complete results and discussion for both short- and long-term outliers detection were discussed in our previous paper [9]. Generally, each of the signals is treated separately to determine the short-term outliers (**Figure 5a**), while long-term outliers are determined using PCA, *i.e.* all signals are treated altogether; thus, the latter indicates malfunctioning in the chemical process (**Figure 5b**). Overall, the signals contained 0.78 – 7.31% of short-term outliers, with an average of 3.27%. In case of long-term outliers, two periods were identified (days 1—13 and 789—790), with 19,837 data points. Only 2 data points from laboratory measurements of V3SS were recorded during the identified long-term outliers. These laboratory measurements were removed from further analysis (total data points: 1,206).

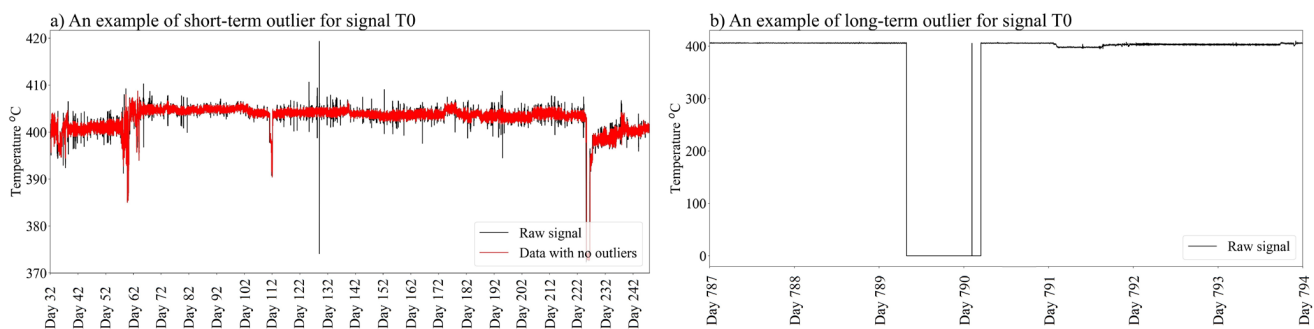

**Figure 5. a)** An example of short-term outliers; **b)** An example of long-term outliers.

At this stage, laboratory measurements and plant sensors are not aligned in time due to different frequencies. The time at which a liquid sample was withdrawn from VDU was used as the ending point to calculate 1-hour average before that measurement. These data were then scaled with zero-mean and unit-variance, and split into training and testing set (70% and 30% - 844 and 362 samples), before training the ANN. The size of the input and output matrices was 1,209 × 31 (samples × features/sensors) and 1,209 × 7 (samples × distillation points), respectively [12].

The minimum error for the testing set during training was achieved at 969 epochs (out of 10,000 epochs), MAE (°C) = 4.64 and 4.76 for the training and testing sets, respectively. The prediction errors (predicted—actual values for V3SS distillation point) are shown in the form of a histogram in **Figure 6a**.

The physicochemical properties of distillation products must lie within a certain range to comply with industrial requirements. Particularly, for some of them, the upper range is detrimental to the quality of the product. Hence, in the case of soft sensors, it is important to understand how many of the predictions, or to what extent, lie outside the specified limits. Nevertheless, specifications for VDU-derived products are difficult to quantify in terms of formal requirements. Instead, in this paper, we analyzed how many predictions lie outside a confidence interval appropriate for a confidence level of 95% (note that visually the errors appear to be normally distributed, as can be seen in **Figure 6a**) [13]. In the trained model, approximately 3% of data points in the testing set lie outside the confidence interval (grey columns in **Figure 6a**). The MAE of the predictions for each distillation point are collated in the table in **Figure 6b**. It can be seen that all values are below the upper SARIMA thresholds determined beforehand, which means that the model's performance is satisfactory for each of the distillation points.

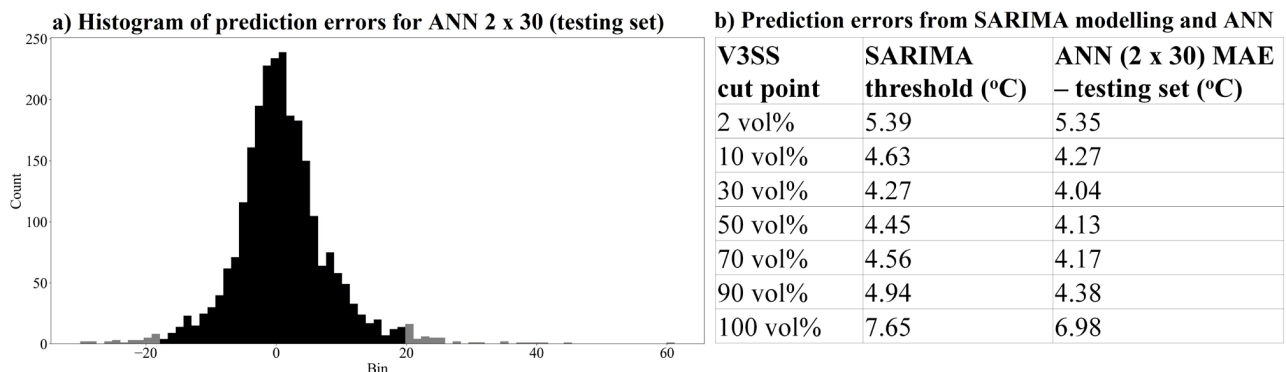

| V3SS cut point | SARIMA threshold (°C) | ANN (2 x 30) MAE – testing set (°C) |
|---|---|---|
| 2 vol% | 5.39 | 5.35 |
| 10 vol% | 4.63 | 4.27 |
| 30 vol% | 4.27 | 4.04 |
| 50 vol% | 4.45 | 4.13 |
| 70 vol% | 4.56 | 4.17 |
| 90 vol% | 4.94 | 4.38 |
| 100 vol% | 7.65 | 6.98 |

**Figure 6. a)** Histogram of deviations in testing set (colors: black - bins inside 95% confidence level, grey - bins outside 95% confidence level); **b)** Comparison of SARIMA thresholds and MAE for the ANN.

The first point of measurements on the 3$^{rd}$ side draw (V3SS line) is the temperature sensor T3 (the VDU diagram was shown previously in **Figure 1)**. It is expected that this temperature has a significant impact on the model predictions. This can be explained by the fact that the sensor T3 measures the draw temperature of V3SS collected at the beginning of the 3$^{rd}$ side draw and distillation curves of the V3SS product (that make up the output of the model).

The feature importance analysis results are shown in **Figure 7** (in the decreasing importance order). It can be seen that T3 has the highest impact on the predictions for all distillation points. T0 has the second highest impact, which can be explained as it measures the feed temperature of the VDU columns (AR), which describes the vapor travelling up the column to liquify into different products, such as V3SS. The sensor T31 measures the return temperature of the reflux which cools down the heavy product in the V3SS stream. Hence, it is expected that sensor T31 and the product quality are correlated.

The VDU process operates under vacuum and so the control of the pressure in the process (PB and PT) is important. Hence, at least one of these pressure measurements should have high potency in the model predictions. As can be seen, sensor PT has a relatively high feature importance value.

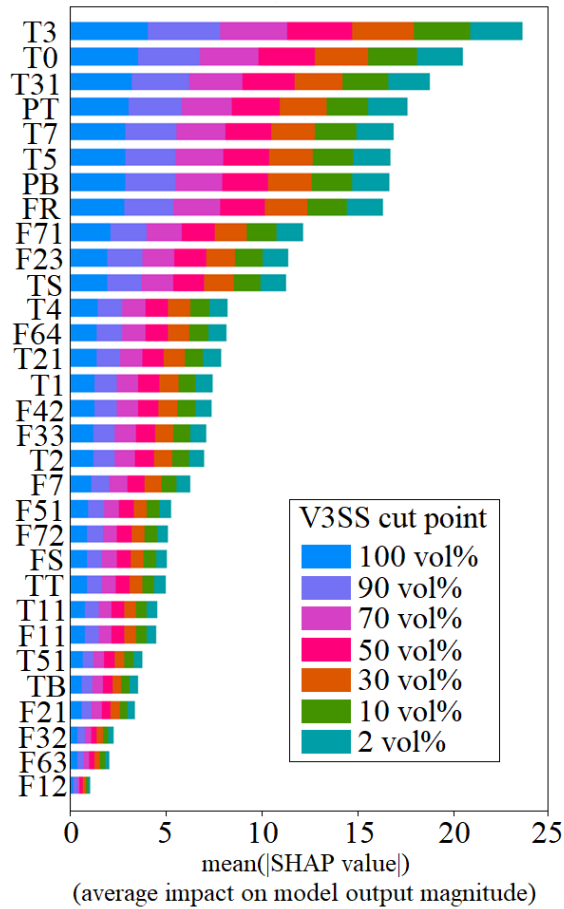

**Figure 7.** Feature importance analysis results (sorted in decreasing order).

**Conclusions**

We presented results of an ANN-based soft sensor model with 2 hidden layers and 30 hidden nodes in each hidden layer. Prior to training the model, a number of pre-processing steps was carried out on the plant sensors data (input for the ANN) and laboratory measurements (output for the ANN) to identify and correct artifacts, such as missing values and outliers.

The primary focus of the ANN was satisfying performance in relation to calculated SARIMA thresholds - *i.e.* the prediction errors of the model below the SARIMA errors. Secondly, we showcased an example of feature importance analysis. This was used to understand how all the plant sensors data (input) influence the prediction of the model which can indicate which sensors are more or less important for the model. This knowledge can support engineers onsite to control the process.

**Acknowledgments**

This work was supported by Innovate UK via a Knowledge Transfer Partnership (No. KTP10916) between the University of Manchester (UK) and Process Integration Limited (UK).